\documentclass[10pt,a4paper]{article}

\usepackage[a4paper,top=20mm,bottom=20mm,left=31mm,right=31mm]{geometry}
\usepackage[T1]{fontenc}
\usepackage[utf8]{inputenc}
\usepackage{lmodern}
\usepackage{microtype}
\usepackage{graphicx}
\usepackage{booktabs}
\usepackage{tabularx}
\usepackage{array}
\usepackage{enumitem}
\usepackage{amsmath}
\usepackage{xcolor}
\usepackage{titlesec}
\usepackage{caption}
\usepackage{float}
\usepackage{placeins}
\usepackage{needspace}
\usepackage{setspace}
\usepackage{fancyhdr}
\usepackage{subcaption}
\usepackage{tikz}
\usetikzlibrary{arrows.meta,positioning}
\usepackage{url}
\usepackage[hidelinks]{hyperref}

\definecolor{Midnight}{HTML}{183B56}
\definecolor{Pale}{HTML}{F1F5F8}
\definecolor{Outline}{HTML}{B7C9D8}
\definecolor{Ink}{HTML}{263343}
\hypersetup{colorlinks=true,linkcolor=Midnight,citecolor=Midnight,urlcolor=Midnight,
  pdftitle={Deep Learning for Deepfake Image Detection: Comparative Model Evaluation and Gradient-Based Visual Attribution},
  pdfauthor={Saad Korchi; Abderrahim WAGA}}
\makeatletter
\renewcommand{\normalsize}{\@setfontsize\normalsize{9.5pt}{11.7pt}}
\makeatother
\setlist[itemize]{leftmargin=1.8em,itemsep=1.5pt,topsep=3pt,parsep=0pt}
\titleformat{\section}{\large\bfseries\color{Midnight}}{\thesection.}{0.65em}{}
\titleformat{\subsection}{\normalsize\bfseries\color{Midnight}}{\thesubsection}{0.6em}{}
\titlespacing*{\section}{0pt}{12pt}{6pt}
\titlespacing*{\subsection}{0pt}{9pt}{4pt}
\fancypagestyle{plain}{%
  \fancyhf{}%
  \fancyfoot[C]{\color{Ink}\footnotesize\thepage}%
}

\begin{document}
\begin{center}
{\fontsize{16.2}{19.0}\selectfont\bfseries\color{Midnight}
Deep Learning for Deepfake Image Detection:\par}
\vspace{1mm}
{\fontsize{15.2}{18.0}\selectfont\bfseries\color{Midnight}
Comparative Model Evaluation and Gradient-Based Visual Attribution\par}
\vspace{3.3mm}
{\fontsize{11.7}{14.0}\selectfont\bfseries Saad Korchi\textsuperscript{*},\quad Abderrahim Waga\par}
\vspace{2mm}
{\fontsize{9.3}{12.0}\selectfont\color{Ink}
Euromed University of Fez, Fez, Morocco\par
\vspace{1.2mm}
\href{mailto:saad.korchi@eidia.ueuromed.org}{\texttt{saad.korchi@eidia.ueuromed.org}},\quad \href{mailto:a.waga@ueuromed.org}{\texttt{a.waga@ueuromed.org}}\par}
\end{center}
\vspace{1mm}
\footnotetext[1]{Corresponding author: \href{mailto:saad.korchi@eidia.ueuromed.org}{\texttt{saad.korchi@eidia.ueuromed.org}}}
\vspace{1.6mm}
\noindent{\color{Outline}\rule{\linewidth}{0.4pt}}
\vspace{1.5mm}

\section*{Abstract}
\noindent
Synthetic facial imagery has made the distinction between authentic and generated content increasingly important for digital forensics and online trust. This study examines an explainable deep-learning workflow for image-level deepfake detection using two contrasting classification strategies: a compact convolutional model trained from scratch and an ImageNet-pretrained MobileNetV2 model adapted through two-stage transfer learning. Experiments were conducted with the balanced 140k Real and Fake Faces dataset, comprising 140,000 facial images. Under the recorded validation protocol, the augmented from-scratch model reached 84.15\% accuracy (loss 0.3625), compared with 95.90\% accuracy (loss 0.1074) for the fine-tuned pretrained model. Gradient-weighted class activation mapping (Grad-CAM) was used to visualize image regions associated with predictions, and a Flask-based prototype was developed for side-by-side inspection. These observations support the practical value of pretrained visual representations in the studied setting while underscoring the need to interpret validation performance alongside dataset provenance, explanation limitations, and evaluation scope. The reported results are validation results, not held-out test estimates.

\noindent\textbf{Keywords:} Synthetic Face Forensics; Transfer Learning; Gradient-Weighted Class Activation Mapping; Model Interpretability.

\section{Introduction}
Improvements in generative modeling have made artificial facial images increasingly convincing. Although synthetic imagery has legitimate applications, the same capability raises concerns about impersonation, fabricated evidence, misinformation, and identity misuse. Reliable media authentication has therefore become relevant to computer vision, information security, and digital forensics.

An automated detector faces more than a binary classification problem. It must also be evaluated in terms of the evidence used to support its decisions, the data distribution under which it was trained, and the conditions in which its predictions may fail. Models that appear accurate on a familiar benchmark can exploit characteristics of its data collection or image-generation pipeline rather than broadly transferable forensic evidence.

This paper studies two complementary deep-learning strategies using the same publicly available balanced facial-image dataset. The first is a compact network learned from scratch; the second uses pretrained visual representations followed by task-specific adaptation. Both models are incorporated into an explainable prototype through Grad-CAM visualizations and side-by-side predictions. The purpose is to document the empirical comparison and to examine how model design, learning dynamics, and interpretability interact in a practical detection workflow.

This study contributes a documented comparison between learning from scratch and transfer learning on a dataset of 140,000 authentic and synthetic facial images. It describes preprocessing, training schedules, fine-tuning procedures, and the recorded validation evidence; integrates Grad-CAM visualizations into an analyst-facing Flask prototype; and carefully distinguishes observed validation findings from claims about generalization or held-out test performance.

\Needspace{12\baselineskip}\section{Related Work}
\subsection{Visual forensic cues and architectural baselines}
Early deepfake detectors relied on local visual inconsistencies and manipulation artifacts. MesoNet illustrated that a relatively compact convolutional architecture can provide useful forgery-detection baselines \cite{mesonet}. Face-warping analysis also showed that geometric artifacts may reveal manipulated facial content \cite{warping}. The FaceForensics++ benchmark subsequently provided a widely used evaluation setting involving multiple manipulation methods and compression conditions, enabling more systematic comparisons \cite{ffpp}.

Modern classifiers can also benefit from pretrained representations. The MobileNetV2 architecture uses inverted residual blocks and depthwise separable convolutions to limit computational cost while retaining expressive visual features \cite{mobilenetv2}. In related computer-vision research, Xception showed the value of depthwise separable convolutions \cite{xception}; ImageNet-scale learning established a broadly used source of initialization for downstream visual tasks \cite{imagenet}. These results motivate comparing a small independently trained detector with a model that reuses pretrained visual features. They do not, however, guarantee that the pretrained model will generalize to previously unseen synthesis methods.

\subsection{Generalization beyond a single manipulation source}
A central challenge is that detectors may learn incidental cues from their training data. Face X-Ray explored blending-boundary features as a route toward broader forgery detection \cite{facexray}, while a survey by Tolosana et al. documented the diversity of face-manipulation methods and their detection challenges \cite{survey}. Shiohara and Yamasaki introduced \emph{self-blended images}, which produce synthetic training examples designed to help a detector learn less manipulation-specific evidence. They evaluated the approach across multiple datasets and manipulation settings \cite{selfblended}. Their findings are particularly relevant to the present study because a classifier trained on a restricted real/synthetic source distribution may not preserve its performance on other generators or alteration mechanisms.

\subsection{Evaluation protocols and newer synthetic-image benchmarks}
Reported accuracy is difficult to compare across studies when preprocessing, splits, training details, and metrics differ. Yan et al. addressed this problem with \emph{DeepfakeBench}, providing a more standardized experimental framework across detectors and datasets \cite{deepfakebench}. Its emphasis on protocol consistency supports the careful reporting of experimental boundaries in this paper, especially the distinction between validation observations and complete held-out test results.

As image-generation methods diversify, benchmark coverage becomes equally important. Zhu et al. introduced \emph{GenImage}, a large-scale benchmark that evaluates generated-image detection under cross-generator and image-degradation conditions \cite{genimage}. Although GenImage is broader than facial deepfakes, its evaluation perspective is directly relevant: good performance on one synthesis distribution does not establish resistance to new generators, blur, compression, or other real-world transformations. These works provide context for interpreting the present comparison as a focused experiment rather than a comprehensive or cross-domain benchmark.

\subsection{Interpretability of visual predictions}
Grad-CAM highlights regions associated with a class prediction by combining gradients with convolutional feature maps \cite{gradcam}. In a forensic workflow, such heatmaps offer an additional diagnostic view of model behavior; they are not causal explanations or evidence that the model has discovered intrinsically reliable forensic features. This distinction guides the interpretation of the visual results reported below.

\Needspace{8\baselineskip}\section{Dataset and Experimental Design}
\subsection{Dataset composition}
Experiments used the public \emph{140k Real and Fake Faces} dataset \cite{kaggle}. According to the project documentation, the authentic class derives from FFHQ-style real face imagery and the synthetic class from StyleGAN-generated faces \cite{stylegan}. The binary classes and the recorded partition sizes are summarized in Table~\ref{tab:dataset}. All reported quantitative comparisons in this manuscript pertain to the documented validation setting.

\begin{table}[htbp]
\centering
\caption{Recorded dataset partition. The existence of a test partition should not be confused with reported test performance.}
\label{tab:dataset}
\renewcommand{\arraystretch}{1.14}
\begin{tabular}{lrrr}
\toprule
\textbf{Partition} & \textbf{Authentic} & \textbf{Synthetic} & \textbf{Total} \\
\midrule
Training & 50,000 & 50,000 & 100,000 \\
Validation & 10,000 & 10,000 & 20,000 \\
Reserved test & 10,000 & 10,000 & 20,000 \\
\midrule
Total & 70,000 & 70,000 & 140,000 \\
\bottomrule
\end{tabular}
\end{table}

\subsection{Preprocessing and augmentation}
Images were resized to $224\times224$ RGB pixels. The from-scratch model used inputs scaled to $[0,1]$, whereas the transfer-learning pipeline applied the standard MobileNetV2 input preprocessing. The final reported scratch-model configuration included random horizontal flips, rotation factor $0.1$, and zoom factor $0.1$ during training. These transformations were not intended to modify the validation inputs.

The different normalization functions follow the requirements of the respective backbones. They also mean that the comparison is between \emph{two complete training strategies}, rather than a controlled substitution of only the final classifier. This distinction matters when interpreting the observed performance gap.

\section{Model Architectures and Training}
\subsection{From-scratch convolutional model}
\begin{figure}[htbp]
\centering
\begin{tikzpicture}[
  scale=1,transform shape,
  every node/.style={font=\scriptsize,align=center},
  stage/.style={draw=Outline,rounded corners=2pt,fill=Pale,text=Ink,minimum width=11.0cm,minimum height=0.65cm,inner sep=3.5pt},
  smallstage/.style={draw=Outline,rounded corners=2pt,fill=white,text=Ink,minimum width=11.0cm,minimum height=0.64cm,inner sep=3pt},
  line/.style={-Latex,semithick,draw=Midnight}
]
\node[smallstage] (input) {\textbf{Input image} \quad $224\times224\times3$ RGB};
\node[stage,below=0.23cm of input] (c1) {\textbf{Block 1} \quad Conv $3\times3$, 32 filters $\rightarrow$ ReLU $\rightarrow$ MaxPool $2\times2$ $\rightarrow$ Dropout 0.25 \\ $112\times112\times32$};
\node[stage,below=0.23cm of c1] (c2) {\textbf{Block 2} \quad Conv $3\times3$, 64 filters $\rightarrow$ ReLU $\rightarrow$ MaxPool $2\times2$ $\rightarrow$ Dropout 0.25 \\ $56\times56\times64$};
\node[stage,below=0.23cm of c2] (c3) {\textbf{Block 3} \quad Conv $3\times3$, 128 filters $\rightarrow$ ReLU $\rightarrow$ MaxPool $2\times2$ $\rightarrow$ Dropout 0.25 \\ $28\times28\times128$};
\node[stage,below=0.23cm of c3] (c4) {\textbf{Block 4} \quad Conv $3\times3$, 256 filters $\rightarrow$ ReLU $\rightarrow$ MaxPool $2\times2$ $\rightarrow$ Dropout 0.25 \\ $14\times14\times256$};
\node[smallstage,below=0.23cm of c4] (gap) {\textbf{Global average pooling} \quad 256-dimensional feature vector};
\node[smallstage,below=0.23cm of gap] (dense) {\textbf{Dense layer} \quad 128 units, ReLU $\rightarrow$ Dropout 0.50};
\node[smallstage,below=0.23cm of dense] (out) {\textbf{Binary output} \quad Dense 1, sigmoid $\rightarrow$ $P(\mathrm{synthetic}\mid x)$};
\foreach \a/\b in {input/c1,c1/c2,c2/c3,c3/c4,c4/gap,gap/dense,dense/out}{\draw[line] (\a.south) -- (\b.north);}
\end{tikzpicture}
\caption{Custom model architecture reconstructed from the recorded layer configuration. Spatial sizes assume same-padded convolutions and $2\times2$ pooling at each of the four stages. The diagram describes the existing model; it is not a newly evaluated architecture.}
\label{fig:architecture}
\end{figure}
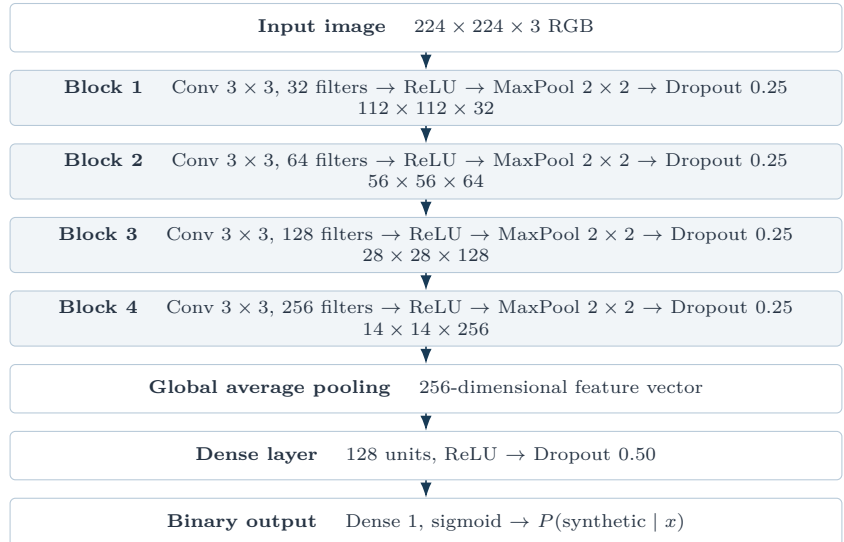

The custom classifier consists of four convolutional blocks. Each block applies a $3\times3$ convolution with ReLU activation and same padding, followed by $2\times2$ max pooling and dropout ($p=0.25$). The filter counts increase through 32, 64, 128, and 256. The feature hierarchy is followed by global average pooling, a fully connected layer with 128 ReLU units, dropout ($p=0.50$), and a sigmoid output for binary classification. The recorded configuration contains approximately 421,000 trainable parameters. Figure~\ref{fig:architecture} makes the layer order and dimensional changes explicit.

\subsection{Transfer-learning model}
The second model uses MobileNetV2 without its original ImageNet classification head. A global-average-pooling stage is followed by a 256-unit ReLU layer, batch normalization, dropout of 0.40, a 128-unit ReLU layer, dropout of 0.30, and a sigmoid output. During the first training phase, the pretrained backbone was frozen while the classification head was optimized for ten epochs at learning rate $10^{-3}$. During the second phase, the final 30 backbone layers were unfrozen for five additional epochs at learning rate $10^{-5}$.

\Needspace{8\baselineskip}\subsection{Optimization protocol}
Both strategies used binary cross-entropy loss and the Adam optimizer. The scratch model was trained over a 15-epoch schedule at batch size 16 and learning rate $10^{-3}$. The documented implementation included mixed-precision computation with a float32 output layer, checkpoint monitoring of validation accuracy, early stopping with patience 4, and learning-rate reduction on plateau with patience 2. The settings are summarized in Table~\ref{tab:training}.

\begin{table}[htbp]
\centering
\caption{Recorded modeling and optimization settings.}
\label{tab:training}
\renewcommand{\arraystretch}{1.14}
\begin{tabularx}{\linewidth}{@{}p{39mm}XX@{}}
\toprule
\textbf{Setting} & \textbf{From scratch} & \textbf{Transfer learning}\\
\midrule
Input size & $224\times224\times3$ & $224\times224\times3$\\
Initialization & He / random & ImageNet pretrained\\
Optimization & Adam; binary cross-entropy & Adam; binary cross-entropy\\
Learning rate & $10^{-3}$ & $10^{-3}$ head; $10^{-5}$ fine-tuning\\
Training stages & 15 epochs & 10 epochs + 5 epochs\\
Output & Sigmoid & Sigmoid\\
\bottomrule
\end{tabularx}
\end{table}

\section{Explainability and Prototype}
\subsection{Grad-CAM visualization}
Grad-CAM was used to identify regions associated with classification outputs. The custom network exposed its last convolutional feature map, while the transfer-learning model used the final spatial feature representation of its backbone. Gradients were aggregated to obtain class-relevant heatmaps, which were then resized and overlaid on the original images.

The visualizations show different patterns of relative emphasis for the same example, with the pretrained model displaying a broader upper-face attribution and the scratch model focusing more sharply near the eyes. The map should not be interpreted as a verified anatomical marker of image manipulation.

\subsection{Web-based model inspection}
\begin{figure}[htbp]
\centering
\begin{subfigure}[c]{.56\linewidth}
  \centering
  \includegraphics[width=\linewidth]{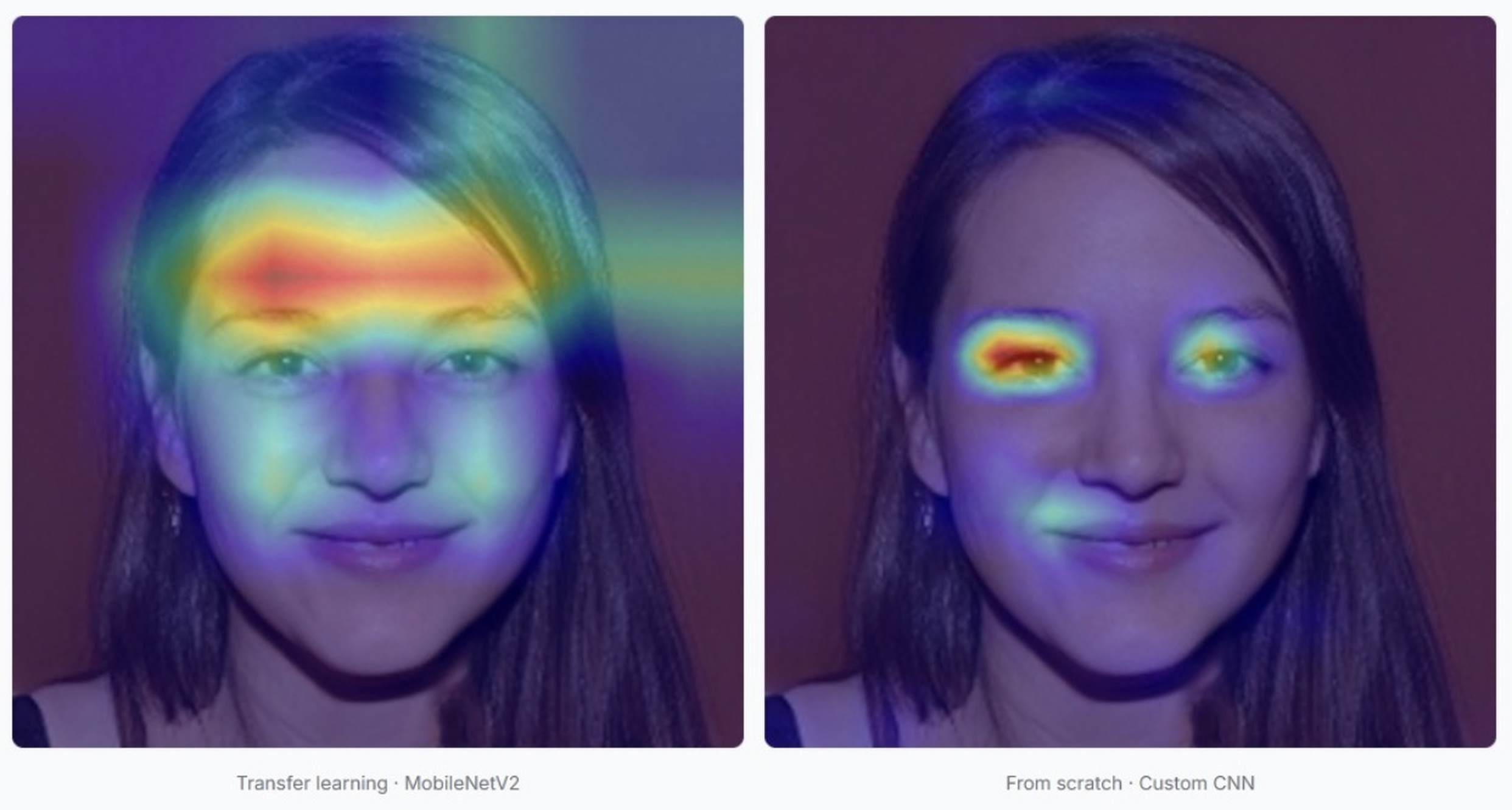}
  \caption{Grad-CAM visual comparison.}
  \label{fig:gradcam}
\end{subfigure}\hfill
\begin{subfigure}[c]{.31\linewidth}
  \centering
  \includegraphics[width=\linewidth]{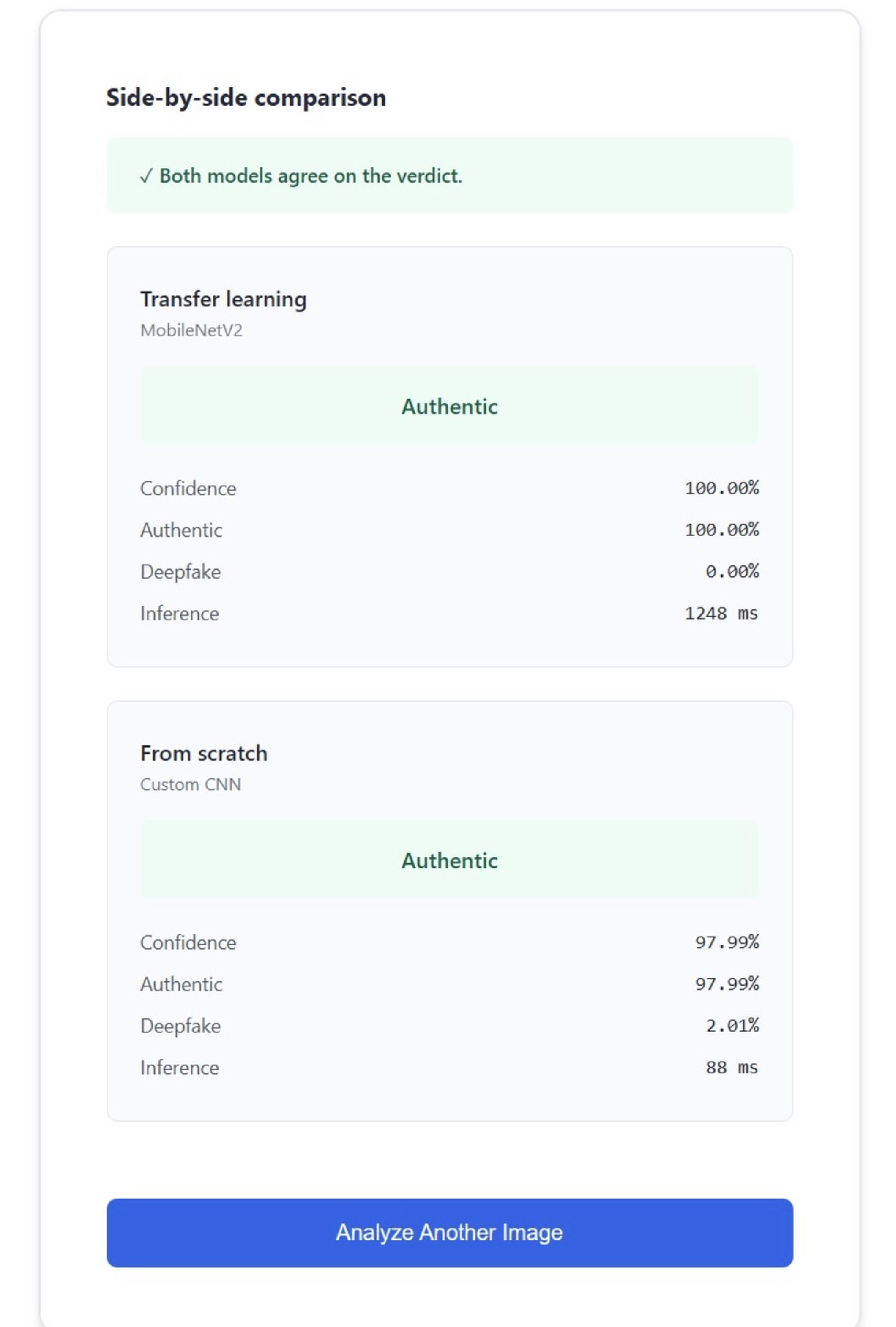}
  \caption{Web inspection interface.}
  \label{fig:interface}
\end{subfigure}
\caption{Qualitative outputs of the experimental workflow. (a) Class-attribution heatmaps for the transfer-learning and custom classifiers on a representative image; spatial attention does not establish causality. (b) Flask prototype displaying side-by-side model predictions, confidence information, and inference times.}
\label{fig:explainability}
\end{figure}

A Flask-based interface was developed to expose the two trained inference paths through a common workflow. It supports model selection, side-by-side predictions, displayed confidence values, and inference-time information. Model weights are cached after loading to simplify repeated use. Figure~\ref{fig:interface} documents one interface state rather than a formal usability or deployment evaluation.

\Needspace{8\baselineskip}\section{Experimental Findings}
\subsection{Learning behavior}
The training curves reflect two different optimization regimes. The scratch model improves progressively during its 15 epochs, whereas the pretrained strategy begins with a stronger representation and subsequently benefits from fine-tuning the final backbone layers. The three preserved curve panels are shown separately to allow the details of the learning trajectories to remain legible.

\begin{figure}[htbp]
\centering
\begin{subfigure}[t]{.46\linewidth}
  \centering\includegraphics[width=\linewidth]{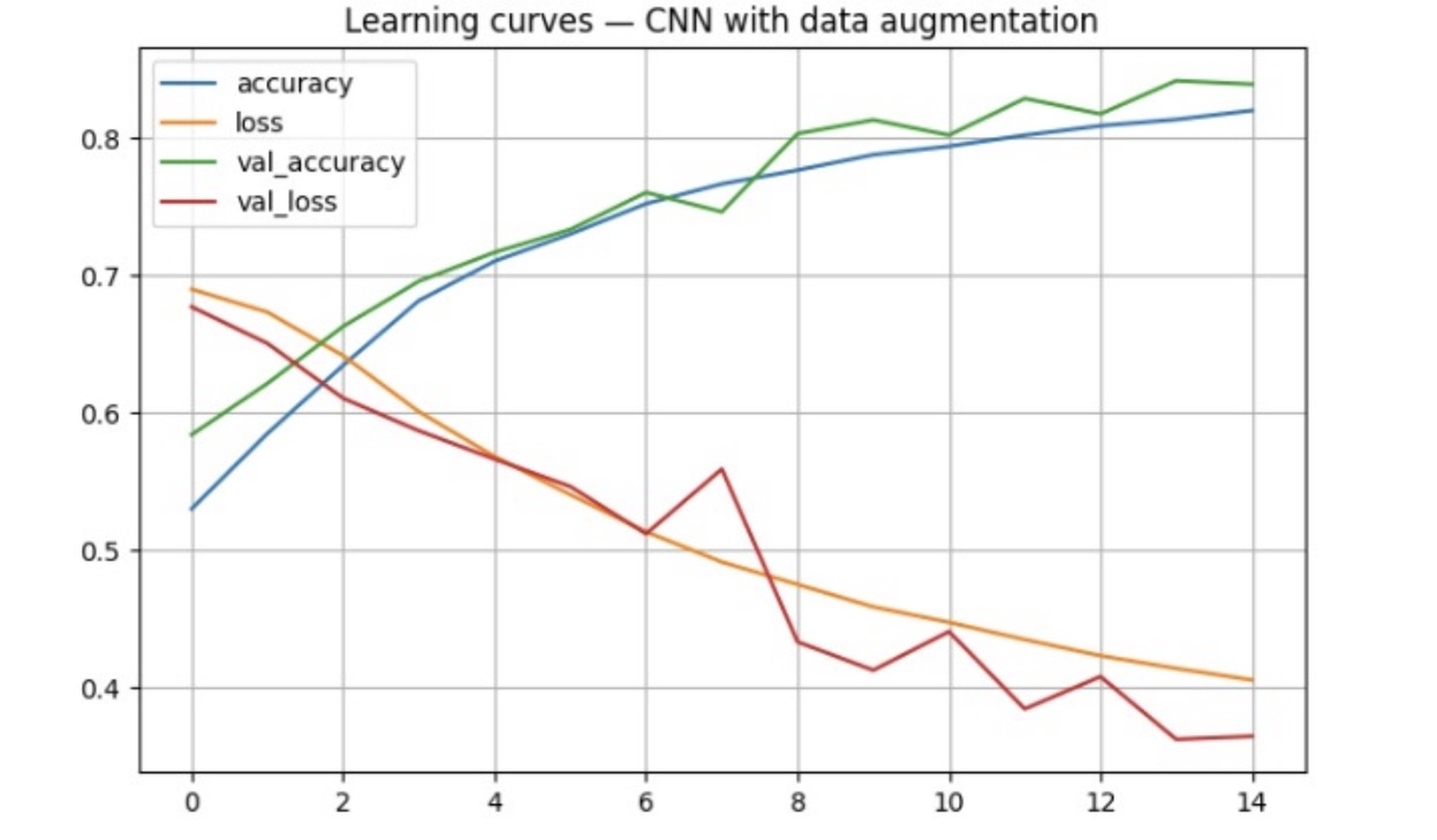}
  \caption{From-scratch training.}
\end{subfigure}\hfill
\begin{subfigure}[t]{.46\linewidth}
  \centering\includegraphics[width=\linewidth]{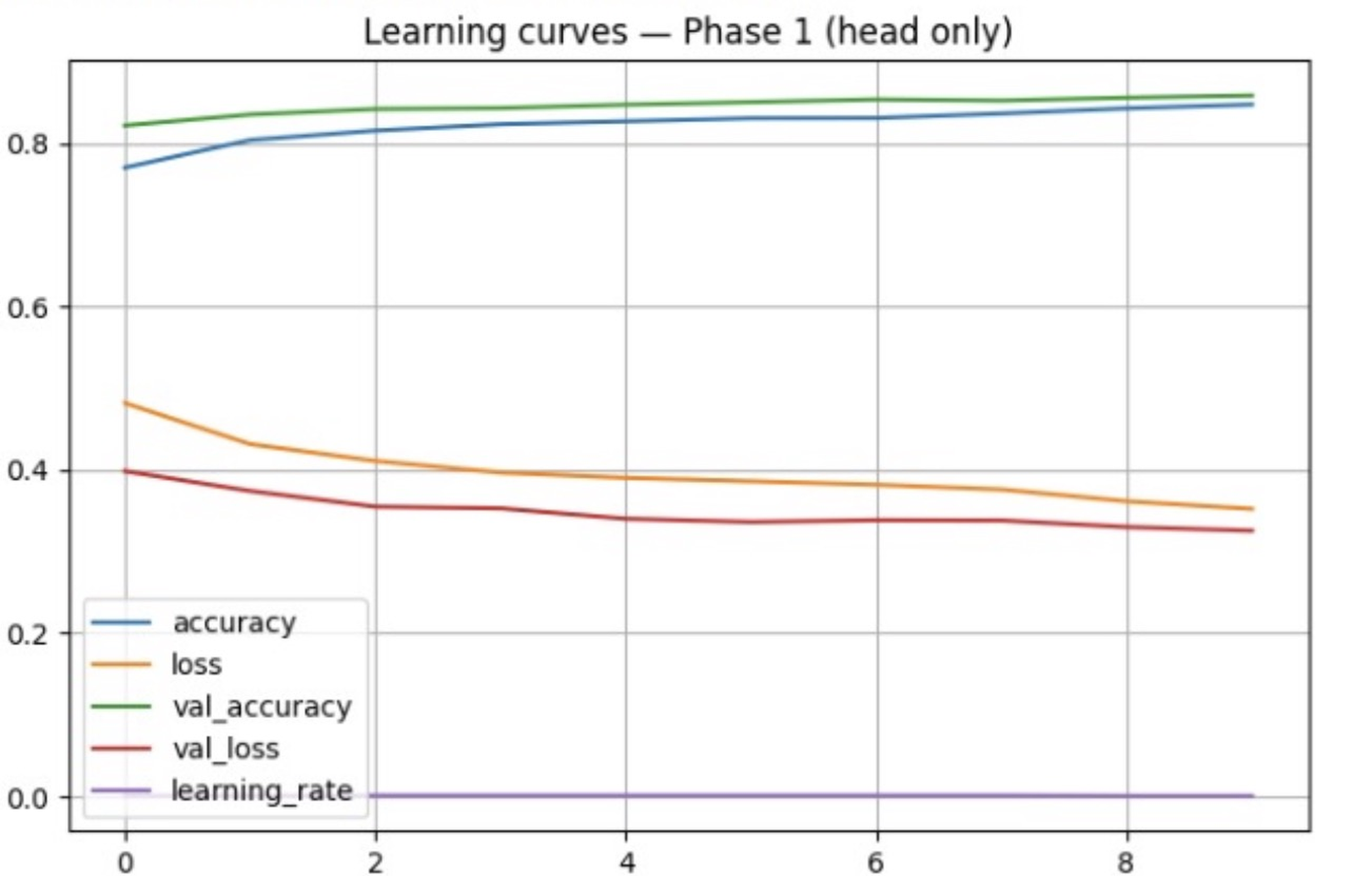}
  \caption{Frozen-backbone phase.}
\end{subfigure}
\vspace{2pt}
\begin{subfigure}[t]{.58\linewidth}
  \centering\includegraphics[width=\linewidth]{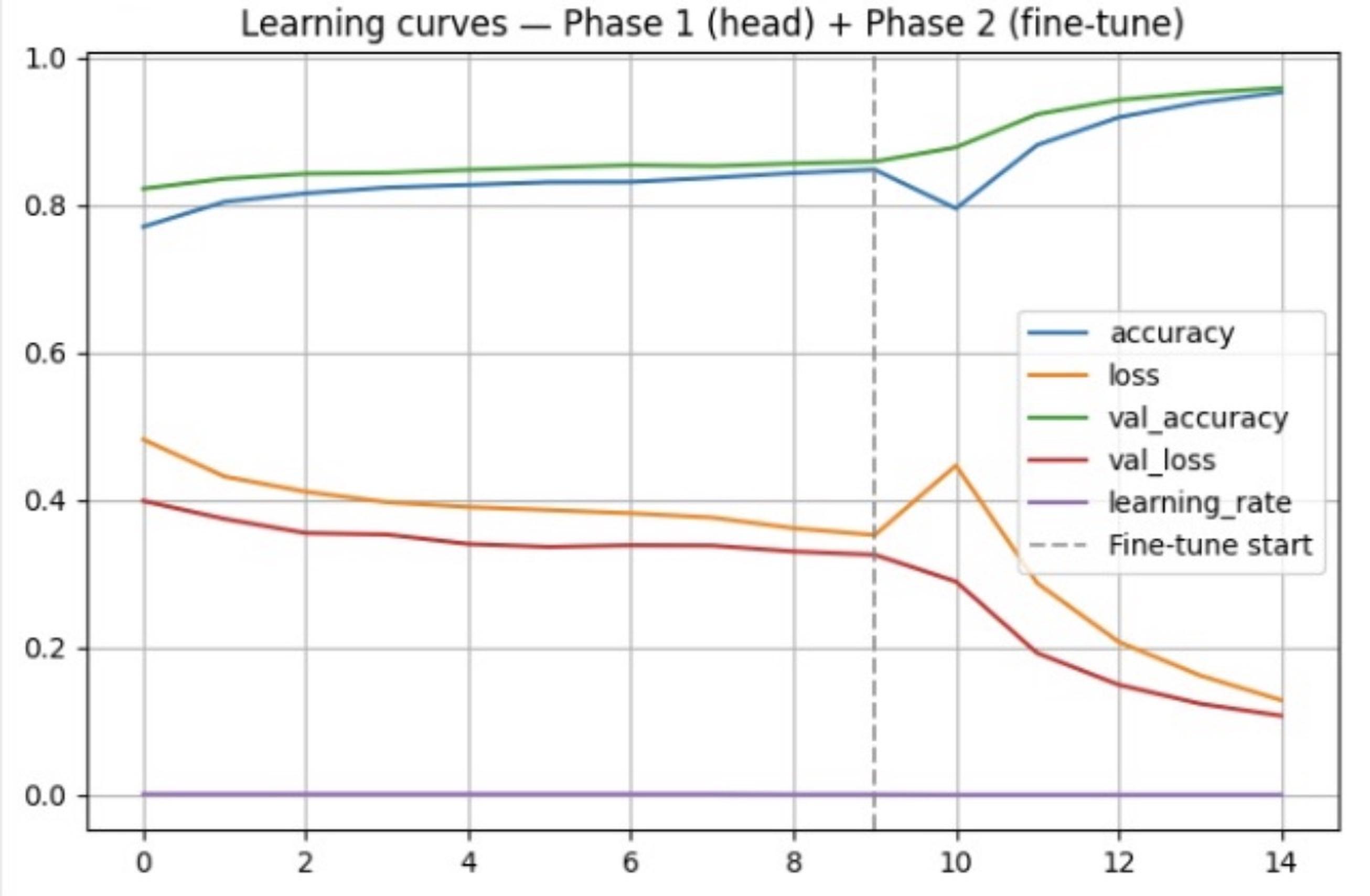}
  \caption{Head training and fine-tuning.}
\end{subfigure}
\caption{Recorded learning curves for the two training strategies. (a) Augmented from-scratch model; (b) frozen-backbone MobileNetV2 training; (c) fine-tuning with the transition marked. All curves reflect the archived experiments.}
\label{fig:curves}
\end{figure}

The final recorded validation accuracy was 84.15\% for the augmented scratch model and 95.90\% for the fine-tuned MobileNetV2 system, an observed gap of 11.75 percentage points under the project configuration. The corresponding validation losses were 0.3625 and 0.1074, respectively. These values are summarized in Table~\ref{tab:results}; they should not be read as test-set performance.

\begin{table}[htbp]
\centering
\caption{Recorded metrics from the final training configurations; all reported evaluation figures are validation figures, not held-out test metrics.}
\label{tab:results}
\renewcommand{\arraystretch}{1.18}
\begin{tabular}{@{}lccc@{}}
\toprule
\textbf{Model} & \textbf{Training acc.} & \textbf{Validation acc.} & \textbf{Validation loss}\\
\midrule
Custom model & 81.33\% & 84.15\% & 0.3625\\
MobileNetV2 & 95.30\% & 95.90\% & 0.1074\\
\bottomrule
\end{tabular}
\end{table}

This comparison is consistent with a practical advantage for pretraining in this setting, but it does not isolate which individual design choice caused the difference. Initialization, backbone capacity, optimization schedules, and preprocessing all differ between the complete pipelines.

\subsection{Representative predictions}
\begin{figure}[htbp]
\centering
\includegraphics[width=.69\linewidth]{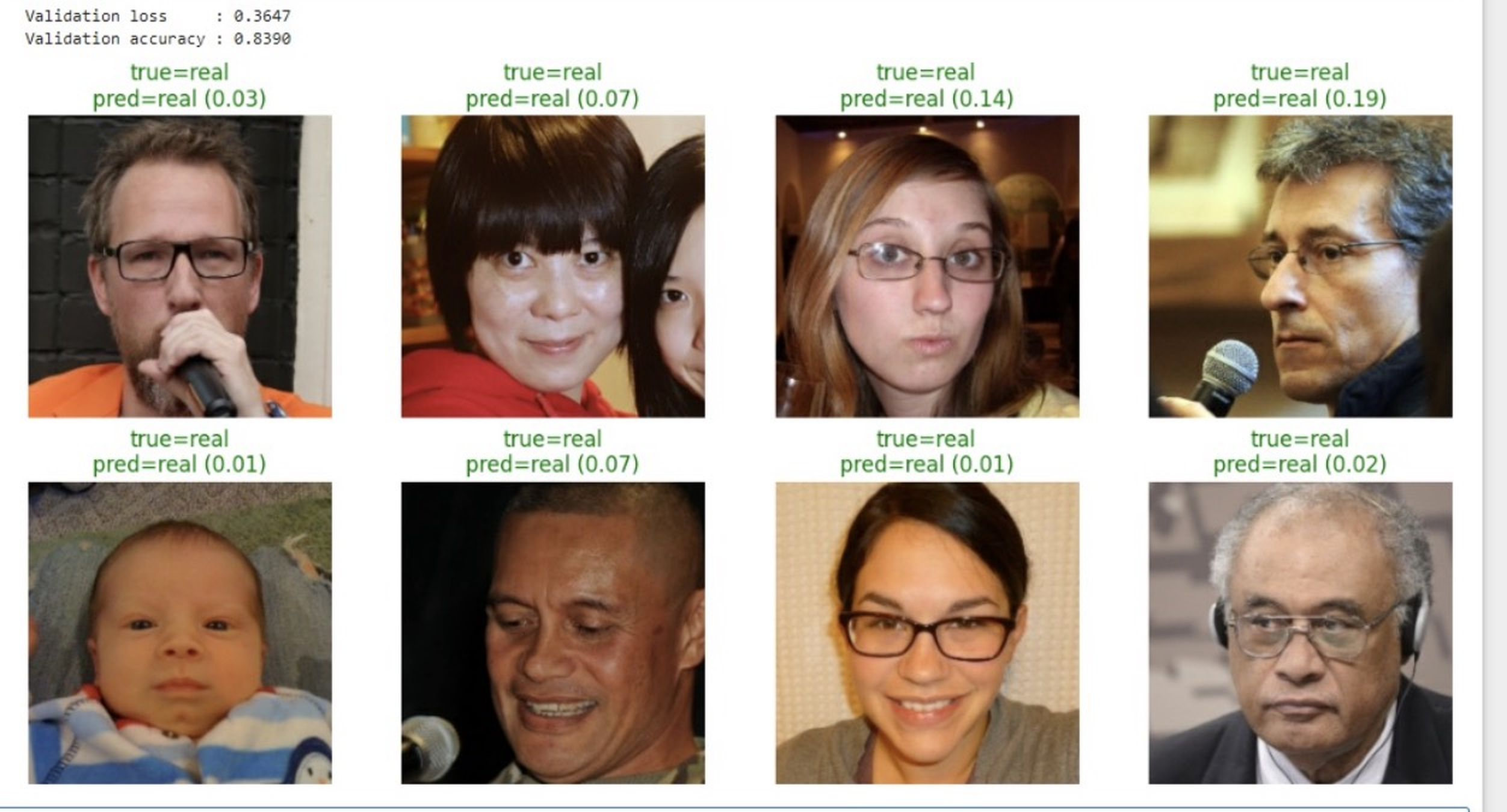}
\caption{Representative validation examples preserved from the original experiments. The illustrated cases are correctly classified authentic faces, not a random or comprehensive sample of errors.}
\label{fig:predictions}
\end{figure}

Figure~\ref{fig:predictions} contains examples retained from the model's validation outputs. The displayed authentic faces were classified correctly, with low synthetic-class probabilities. Such examples help verify the inference workflow qualitatively but do not measure sensitivity to diverse forgeries or characterize the model's error distribution.

\FloatBarrier
\section{Discussion}
The validation results suggest that transferred representations offered a substantial benefit over learning all features from random initialization for this particular dataset and training schedule. The custom model remains a useful lightweight baseline: its architecture is explicit and relatively compact, and it makes the cost of learning visual features from scratch observable within the same research project.

Interpretability adds context but does not by itself establish trustworthiness. A Grad-CAM map can help an analyst investigate where a model's prediction is spatially sensitive; it cannot certify that the highlighted region is the genuine cause of an alteration. The two models can also produce similar decisions from different visual cues. Such discrepancies deserve additional testing rather than an automatic judgment that one explanation is superior.

More broadly, the benchmark used here is well suited to examining a specific real-versus-StyleGAN classification setting, but that setting is narrower than the variety of synthetic content encountered in practice. The literature on self-blended images, unified benchmarking, and cross-generator evaluation \cite{selfblended,deepfakebench,genimage} illustrates why future experiments should explicitly include novel synthesis techniques, image degradation, and testing on independent datasets.

\section{Limitations and Future Work}
Several limitations shape how these findings should be used. First, the reported numerical results originate from validation outputs preserved during the project. Although the dataset includes a reserved test partition of 20,000 images, a complete export of held-out test accuracy, precision, recall, F1-score, ROC-AUC, and confusion-matrix results was not available for this manuscript. Reporting the observed validation advantage is appropriate; describing it as demonstrated performance on unseen test data would not be.

Second, the experiment compares two complete learning pipelines under one main split and training schedule. Their architectural complexity, pretraining, preprocessing, and optimization procedures are different. Multiple random seeds and matched experimental controls would be needed to assess run-to-run variability and identify which factors contribute most strongly to the observed gap.

Third, the evaluation dataset represents a limited set of image sources. A classifier may exploit subtle distinctions between the original real-image collection and its synthetic-image generator. Such correlations can lead to optimistic in-distribution results while providing little protection against diffusion-generated imagery, face swaps, image recompression, or other transformations not included in the study.

Finally, the attribution examples and prototype demonstrate useful functionality but have not been subjected to a controlled assessment of explanation reliability, human analyst decisions, or robustness against deliberate perturbations. Next steps should therefore prioritize evaluation of frozen checkpoints on the untouched test partition, complete reporting of classification and calibration metrics, cross-generator experiments, additional training seeds, and stronger verification of explanatory maps. These extensions would help determine how far the current findings carry beyond this initial experimental setting.

\section{Research Ethics and Reproducibility}
The study uses an existing publicly accessible image dataset and did not involve new direct recruitment of human participants. Facial datasets nevertheless raise questions of consent, provenance, licensing, and downstream misuse. Any reuse or release of data and illustrative images should respect the applicable dataset terms and privacy obligations. Model outputs should support, rather than replace, independent verification and human judgment in consequential forensic settings.

Implementation was carried out in Python using TensorFlow/Keras, OpenCV, and Flask. The recorded GPU environment was an NVIDIA RTX 4070 under WSL2 Linux. Input size, loss function, optimization parameters, and available figures are described above to permit technical scrutiny. A complete reproducibility release would additionally include versioned code and environment specifications, fixed random seeds, model checkpoints where permitted, and exported predictions on the reserved test split.

\section{Conclusion}
This paper documented an image-level deepfake detection experiment combining two deep-learning strategies with visual explanation and a practical inspection interface. Within the recorded validation conditions, the fine-tuned pretrained model reached 95.90\% accuracy compared with 84.15\% for the augmented scratch model. The comparison provides evidence of the usefulness of transfer learning in this configuration, but not of generalization to unseen generators or real-world forensic deployments. The contribution of the work lies in its transparent experimental comparison and its integration of predictive results with visual model inspection. Extending the evaluation beyond this dataset remains the next scientific priority.

\section*{Data and Code Availability}
The image dataset is publicly available as \emph{140k Real and Fake Faces} \cite{kaggle}. The present manuscript describes recorded methodology and figures from the project. No public code repository or held-out test metrics are claimed for this version.

\section*{Acknowledgments}
The authors would like to express their sincere gratitude to the Euro-Mediterranean University of Fez (UEMF) for the academic environment, computational resources, and institutional support that enabled and facilitated this research work.

\Needspace{10\baselineskip}
{\setstretch{1.05}\footnotesize
\bibliographystyle{unsrt}
\bibliography{references}
}
\end{document}